# Multiclass Classification of Cervical Cancer Tissues by Hidden Markov Model

Sabyasachi Mukhopadhyay*, Sanket Nandan*; Indrajit Kurmi**
*Indian Institute of Science Education and Research Kolkata
**Indian Institute of Technology Kanpur.

**Abstract**

In this paper, we report a hidden Markov model based multiclass classification of cervical cancer tissues. This model has been validated directly over time series generated by the medium refractive index fluctuations extracted from differential interference contrast images of healthy and different stages of cancer tissues. The method shows promising results for multiclass classification with higher accuracy.

## Introduction

Cancer diagnosis through optical means is a hot area of current research. There are efficient signal processing tools like wavelets, multifractal detrended fluctuation analysis (MFDFA) which have shown promising outcomes in case of disease diagnosis [2, 3, 4]. We would like to note that these linear techniques are limited for binary classifications.

The current research works are more focused on probabilistic classifiers for extracting the hidden features in biological tissues and enhancing the accuracy of classification. The probabilistic classifiers like support vector machine (SVM), relevance vector machine (RVM), minimum distance classifier like Mahalanobis distance were found to have a lot of applications in biomedical domain for cancer diagnosis purpose [5, 6, 8, 11]. In this paper, hidden Markov model (HMM), which is a dynamic Bayesian model, has been used for classifying cervical tissue samples of healthy and different stages of cancer.

## Theory

Suppose we have a non-empty finite element set of observation sequence, given by

$$O = (O_1, O_2, \ldots\ldots, O_\tau);$$

and $\Omega$ be the outcome (sample) space i.e. set of all possible outcomes or observations such that $|\Omega| > \infty$, thereby $O_t \in \Omega \forall t = 1, 2, ......, \tau$.

Assume there exists some hidden state sequence, $\{X_t\}_{t=1,....,\tau}$, which follows a Markov process, associated with the observation (data) sequence such that

$$X_t O = O_t \forall t; \text{ for some } O. \qquad (1)$$

Take $P = (p_{ij})_{n \times n}$ to be the transition probability matrix for the underlying Markov process with finite, discrete state space $X = \{1, 2, ....., n\}$; so that $p_{ij} := P(X_{t+1} = j | X_t = i)$

The structure of the matrix, $O$, is given by, $O = \{o_j(k)\}_{n \times |\Omega|}$, commonly known as the observation probability matrix, such that $o_j(k) := P(O_t = k | X_t = j)$.

This model $\theta = (P, O, \pi)$, where $\pi$ is the initial state (probability) distribution, is known as hidden Markov Model.

**Experimental Method**

The refractive index in space (two dimensional area of a tissue section) of the cervical tis-sue section with thickness $5\mu m$ and lateral dimension $\sim 4mm \times 6mm$, is the required stochastic (observed through measurement) variable for our study. The unstained tissue section has been concerned with the standard methods of tissue dehydration, thereafter embedding in wax, then sectioning under a rotary microtome and final dewaxing, performed on glass slides. The diferential interference contrast (DIC) microscope (Olympus IX81, USA) measures the distribution of the refractive index on the section surface area of the tissue section. Separately, a CCD camera (ORCA-ERG, Hamamatsu) with pixel dimension $6.45\mu m$ and a point spread function microscope with width $\sim 0.36\mu m$ records the images of the stromal regions with 60X magnification. The multiclass classification has been performed over pixel-wise horizontally unfolded data. The DIC sample images are shown in fig.1. After unfolding the DIC images in a horizontal direction, it has been treated with Normal Testing method in the following ways.
1. Fluctuations= (Data-mean)/standard deviation

2. Computing the cumulative sum of fluctuations to generate the time series.

The above two steps are performed in order to process the medium refractive index fluctuations embedded in DIC images by HMM.

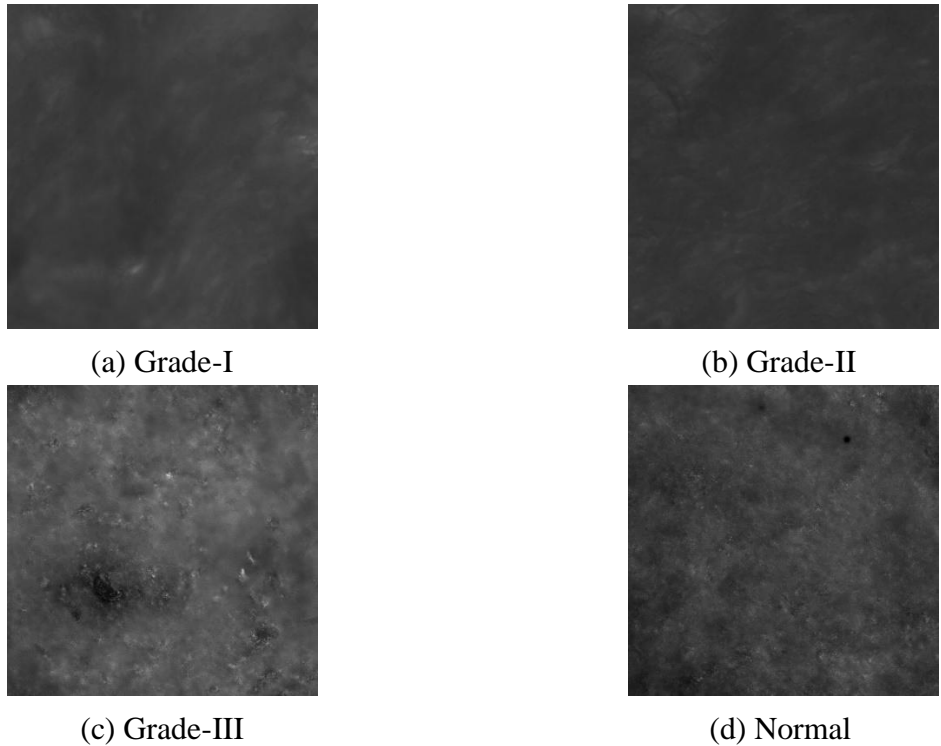

(a) Grade-I  (b) Grade-II

(c) Grade-III  (d) Normal

Figure 1: DIC images of different stages (Grade-I,II,III) of cancer and healthy (normal) tissues respectively.

**Results and Discussions**

Considering a tissue with potentially four different stages, namely, Grade I, II, III and normal (healthy), the horizontal surface of the plot has been partitioned into $4\times 4$ rectangles, where prediction of each stages can lead to inaccurate detection of some other stage if there exists some non-zero percentage of detection in the off-diagonal rectangles given a particular stage to measure. The height or the perpendicular to the surface, z-axis displays the percentage of prediction of a given stage for the corresponding class (horizontal rectangles). The graph fig.2 shows clear distinction among the four different stages of the tissue. Data from normal cells are accurately predicted (with probability 1) as normal cells all the time creates a zero probability of false prediction in the rectangles (classes) of Grade I, II and III. The graph also shows that neither Grade I, nor II, nor III class is predicted as normal stage and hence it presents an advantage of classifying Grade I, II and III together, i.e., cancerous cells and normal cells. Similarly Grade I and II stages are also predicted in the correct class with 100% accuracy. But it is evident from the graph that there is an error in the correct prediction of Grade III stage. Grade III cells are correctly predicted with 97.34% accuracy and error in prediction of Grade III stage is 2.66% that falls in the class of Grade II. This error in prediction implies wrong prediction of a Grade III cell as Grade II cell with 2.66%, but the results never predict it as a Grade I or normal cell.

For the purpose of experimentation, we first train a 17 model states in the hidden Markov model for each of the categories (stages). The training set for each category includes time series data obtained from experimentation. The model trained by the training data is defined as $\theta = (P, O, \pi)$ with state space $S = \{s_1, s_2, \ldots\ldots\ldots, s_{17}\}$. $\pi$ denotes the prior probabilities, elements of $P$ are the transition probabilities and elements of $O$ denote the emission prob-abilities. Prior probabilities are first selected as a random function. $P$ and $O$ are modelled as Gaussian densities with mean 0 and variance 1. Then the data is trained on the model iteratively to fit and modify the model using EM (Expectation maximization) algorithm. The model is optimized using Lagrange multipliers. We use forward and backward algorithm to compute a set of sufficient statistics for our EM step tractably. Once the model is sufficiently trained for a given sequence of data we calculate the likelihood of sequence with model for each category, i.e., we calculate $P\left(\dfrac{X}{\theta_i}\right)$ which is the sum of the joint likelihoods of the sequence over all possible state sequences allowed by the model for each category.

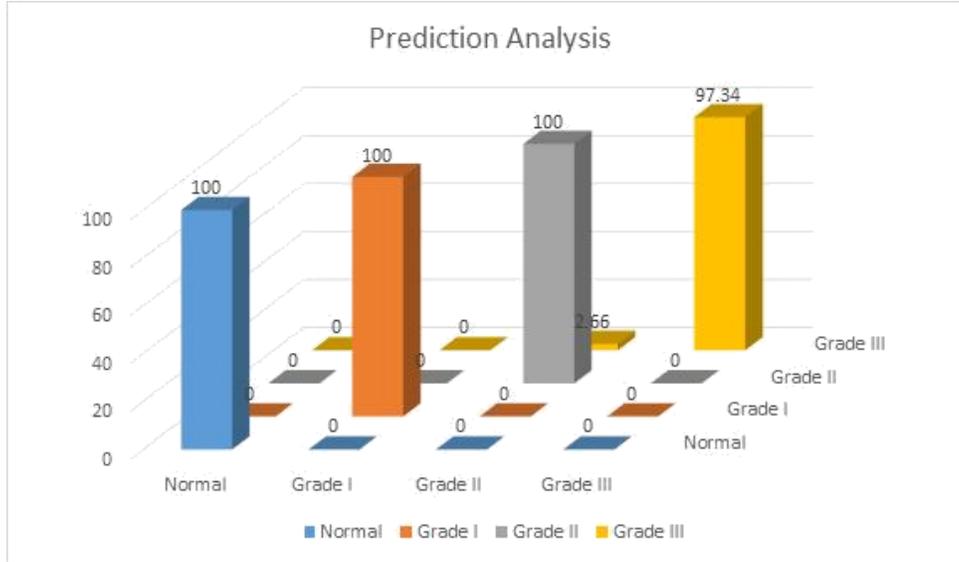

Figure 2: HMM based cervical tissue classification among healthy and different stages of cancer.

**Conclusion**
In conclusion, use of HMM has been found to be quite effective in multiclass classification purpose among the normal and different stages of cancer. Authors hope that the current study of HMM based application in DIC images of human cervical tissues will help the researchers to move this field forward.

**Acknowledgment**
The authors thank Dr. Asha Agarwal, G.S.V.M. Medical College and Hospital, Kanpur, for provision of the histopathologically characterized tissue samples.